\documentclass[letterpaper, 10 pt, conference]{ieeeconf}  
\IEEEoverridecommandlockouts
\usepackage[letterpaper, left=0.75in, right=0.75in, bottom=0.75in, top=1in]{geometry}
\usepackage{graphics} 
\usepackage{mathptmx} 
\usepackage{amsmath}  
\usepackage{amssymb}  
\usepackage[export]{adjustbox} 
\usepackage{bm}
\usepackage{subcaption} 
\usepackage{tikz}
\usepackage[compact]{titlesec}
\usepackage{stfloats}
\usepackage{xltabular}

\newcommand{\etal}{\textit{et al.~}}

\makeatletter
 \let\NAT@parse\undefined
 
 \newcommand{\Rmnum}[1]{\expandafter\@slowromancap\romannumeral #1@}
 
\makeatother
\usepackage[pdftex,
        colorlinks=true,
        bookmarks=true,
        citecolor=red,
        linkcolor=blue,
        urlcolor=black,
        pagebackref=false,
        pdfstartview=Fit,
        breaklinks=true
            ]{hyperref}
\usepackage{cleveref}
\let\labelindent\relax
\usepackage{enumitem}

\title{\LARGE\bf
Whole-Body Control for Velocity-Controlled Mobile Collaborative Robots Using Coupling Dynamic Movement Primitives
}

\author{Zhangjie Tu$^{1}$, Tianwei Zhang$^{1}$, Lei Yan$^{2}$, Tin lun Lam$^{1,3,\dag}$
\vspace{-2cm}
\thanks{This work is supported by the funding AC01202101103 from the Shenzhen Institute of Artificial Intelligence and Robotics for Society.}
\thanks{$^{1}$Shenzhen Institute of Artificial Intelligence and Robotics for Society (AIRS), Shenzhen.}%
\thanks{$^{2}$School of Mechanical Engineering and Automation, Harbin Institute of Technology, Shenzhen.}
\thanks{$^{3}$School of Science and Engineering, The Chinese University of Hong Kong, Shenzhen.
}
\thanks{$^{\dag}$Corresponding author: {\tt\small tllam@cuhk.edu.cn}}%
}

\begin{document}
\maketitle
\titlespacing{\section}{0pt}{2ex}{1ex}

\begin{abstract}
In this paper, we propose a unified whole-body control framework for velocity-controlled mobile collaborative robots which can distribute task motion into the arm and mobile base according to specific task requirements by adjusting weighting factors. Our framework focuses on addressing two challenging issues in whole-body coordination: 1) different dynamic characteristics of the mobile base and the arm; 2) avoidance of violating both safety and configuration constraints.
In addition, our controller involves Coupling Dynamic Movement Primitive to enable the essential capabilities for collaboration and interaction applications, such as obstacle avoidance, human teaching, and compliance control. 
Based on these, we design an adaptive motion mode for intuitive physical human-robot interaction through adjusting the weighting factors.
The proposed controller is in closed-form and thus quite computationally efficient. 
Several typical experiments carried out on a real mobile collaborative robot validate the effectiveness of the proposed controller.
\end{abstract}
\section{Introduction}
Due to the fast development of collaborative robot and autonomous mobile robot, integrated Mobile Collaborative Robots (MCR) are increasingly applied in a wide range of industrial and tertiary scenes. 
Compared with fixed-base robots, MCR offer various benefits including higher redundancy and wider working range \cite{zhang_Decomposed_2021}. At the same time, they bring about some inherent challenges from coordination (between the arm and base), constraints avoidance (both safety and configuration), and intuitive interaction. 
The first challenge is how to coordinate and integrate the different dynamics and motion accuracy of the base and arm into a whole-body controller (WBC). 
Generally speaking, {most of industrial MCR commonly uses the velocity-controlled interface while its WBC is designed based on the weighted pseudo-inverse approach \cite{li_design_2020}, \cite{jia_coordinated_2014}.
However, the issue of different bandwidths between the arm and base, which causes the vibration of the end-effector \cite{yamamoto_compensation_1994}, remains unresolved}.

As safety is the highest priority for physical Human-Robot Interaction (pHRI), MCR are expected to have compliant behavior and collision avoidance capability when it is in-motion. In practice, it often happens that people or moving obstacles push the robot to converge to its singularity or joint limits, where the robot loses the ability of performing Cartesian space tasks. However, it is still very challenging to design a multi-hierarchy control strategy that allows the robot to deviate from the planned trajectory to satisfy the higher priority constraints like avoiding joint limits. 





In addition to collaboration, intuitive pHRI is further demanded for MCR. For instance, more arm motion is expected when collaborating in local and narrow space while more base motion for large range displacements. However, how to assign direct physical accessibility for human and reasonable workspace to MCR remains an open question.




To address the above challenges, we first propose a Coupling Dynamic Movement Primitive (CDMP), and further design a novel unified whole-body controller (\textit{CDMP-WBC}) based on CDMP for velocity-controlled MCR. This work contributes to:
\begin{enumerate}
	\item a unified whole-body controller for velocity-controlled MCR. It distributes the motion between the arm and base, and contains a motion mode of smooth pHRI based on adjustable weighting matrix.
	\item a robust whole-body coordination scheme that not only deals with the different dynamics of base and arm, but also manages multi-task priorities of tracking trajectories and avoiding various constraints including obstacles, singularity, and joint limits.
	\item we divide CDMP as Stiffness-coupled, Admittance-coupled, and propose a new hybrid Stiffness/Force Control type, by which the proposed controller can handle obstacle avoidance, human teaching, and hybrid Stiffness/Force control in a unified way.
\end{enumerate}


\section{Related Work}
%
%
Most of the current research work on whole-body control is based on torque-controlled interface, such as \cite{yamamoto_compensation_1994}, \cite{wu_passive_2021}, \cite{wu_unified_2021}. {Due to the high price of torque-controlled robots today, these methods are not common in industrial applications.}
For velocity-based whole-body controllers commonly used in industrial scenarios, \cite{jia_coordinated_2014}, \cite{xing_enhancement_2020} improves motion accuracy by assigning motion to the arm as much as possible, without directly dealing with the different dynamic characteristics and control bandwidths of the base and arm. 
In addition, \cite{li_design_2020} proposes a whole-body hierarchical controller based on null-space projection, also without considering the different dynamic characteristics of the base and arm. 

Heins \etal \cite{heins_mobile_2021} proposed a WBC for velocity-controlled industrial mobile manipulators based on whole-body motion distribution strategy with QP optimization. It considers constraints of singularity and obstacle avoidance, and allows the end-effector to deviate from the desired trajectory when obstacle avoidance conflicts with task motion, which improves the robustness of WBC, but also does not consider the different dynamic characteristics of base and arm, and this type of optimization-based approach is relatively computationally intensive.
Similarly, \cite{pankert_perceptive_2020} proposes an MPC-based controller using the SLQ optimization algorithm for trajectory tracking, while considering mechanical stability, joint limit, obstacle avoidance and other constraints. It has higher complexity and longer solution time than QP-based algorithm.

To achieve intuitive pHRI, \cite{navarro_framework_2017} first proposes an adaptive strategy based on weighting factors for motion assignment, but does not consider the different dynamical characteristics of base and arm, and has no function of obstacle avoidance.

\section{Preliminaries}
\subsection{Kinematic Modeling}
\begin{figure}[tbp]
    \centering
    \includegraphics[width=0.75\linewidth]{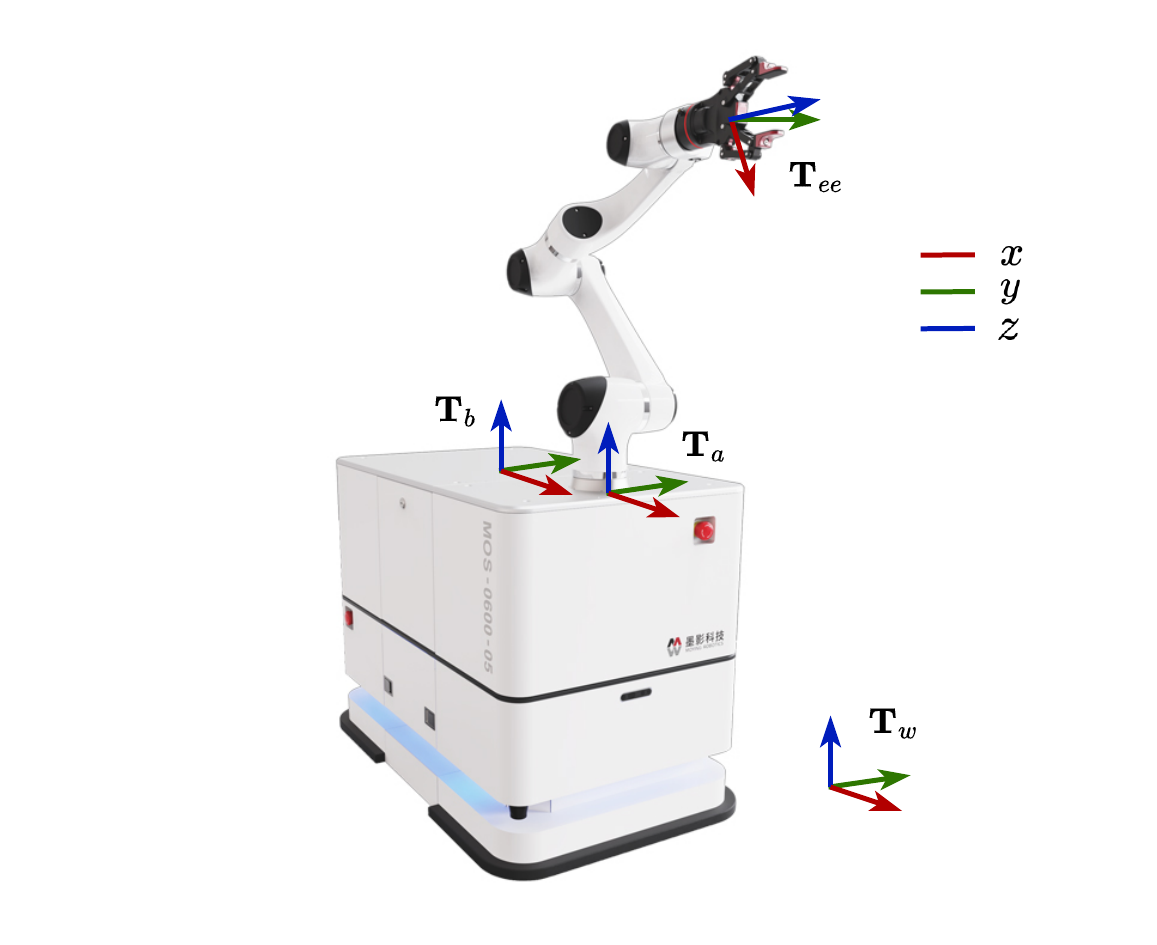}
    \caption{The velocity-controlled MCR and reference frames. $T_w$, $T_a$, $T_b$, and $T_{ee}$ denote the reference frame of the world, the arm, the mobile base, and the end-effector respectively.}
    \label{fig:me5_frames}
\end{figure}

As shown in Fig. \ref{fig:me5_frames}, the MCR is composed of a 3-DoF velocity-controlled omnidirectional wheeled mobile base and a 6-DoF position-controlled manipulator with a wrist-mounted 6-D Force/Torque sensor.
The configuration of the 9-DoF robotic system can be described by $\bm{q}_{wb} = [\bm{q}^T_{b}, \bm{q}^T_{a}]^T$, where $\bm{q}_{b} = [x_b, y_b, \phi_b]^T$ represents the postion and yaw angle of the mobile base w.r.t $T_w$, $\bm{q}_{a} = [\theta_1, \theta_2, ..., \theta_6]^T$ represents the joint angles of the arm. As the velocity command can be integrated into position command, we choose velocity control inputs as motion command to the arm for the consistency with the mobile base. $\bm{V}_{b} = [v^x_b, v^y_b, \omega^z_b]^T \in \mathbb{R}^{3}$ denotes the Cartesian velocity control inputs of the mobile base w.r.t its body frame $T_b$ and $\bm{V}_a = [\bm{v}^T_a, \bm{\omega}^T_a]^T \in \mathbb{R}^{6}$ denotes the arm's w.r.t $T_a$, thus the complete velocity input to the MCR is $\bm{V}_{wb} = [\bm{V}^T_{a}, \bm{V}^T_b]^T \in \mathbb{R}^{9}$.
Then, the end-effector's reference velocity w.r.t $T_b$ can be calculated as
\begin{equation} \label{eqn:fk}
\begin{split}
	\bm{V}_{ee} &= \begin{bmatrix}
	\bm{v}_{ee} \\
	\bm{\omega}_{ee}
	\end{bmatrix} = \begin{bmatrix}
	\bm{R}_{ba} \cdot \bm{J}_a & \bm{J}_b \cdot ^b\bm{R}_w
	\end{bmatrix}
	\begin{bmatrix}
	\dot{\bm{q}}_a \\
	\dot{\bm{q}}_b
	\end{bmatrix} = ^w\bm{J}_{wb} \cdot \dot{\bm{q}}_{wb}\\
	&= \begin{bmatrix}
	\bm{R}_{ba} & \bm{J}_b
	\end{bmatrix}
	\begin{bmatrix}
	\bm{V}_a \\ \bm{V}_b
	\end{bmatrix} = \bm{J}_{wb} \cdot \bm{V}_{wb}
\end{split}
\end{equation}
where $\bm{R}_{ba} = 
\begin{bmatrix}
^b\bm{R}_a & 0 \\
0 & ^b\bm{R}_a
\end{bmatrix} \in \mathbb{R}^{6\times6}$; $^b\bm{R}_a \in SO(3)$ is the constant matrix transforming a vector from $T_a$ to $T_b$ which is usually an identity matrix; $^b\bm{R}_w = Rot(\bm{\omega}_b, -\phi_b) \in SO(3)$ is the transformation matrix from $T_w$ to $T_b$;
 $\bm{J}_a \in \mathbb{R}^{6\times6}$ is the geometric Jacobian of the arm w.r.t $T_a$, while $\bm{J}_b \in \mathbb{R}^{6\times3}$ is the base's w.r.t $T_b$ and 
\begin{equation}
\bm{J}_b = \begin{bmatrix}
1 & 0 & -^by_{ee}\\
0 & 1 & ^bx_{ee} \\
0 & 0 & 0 \\
0 & 0 & 0\\
0 & 0 & 0\\
0 & 0 & 1\\
\end{bmatrix}
\end{equation}
where $^bx_{ee}$ and $^by_{ee}$ are the end-effector's coordinates along the $x$ and $y$ axes of $T_b$. $\bm{V}_{ee}$ can be calculated by
\begin{equation}
\bm{V}_{ee} = \begin{bmatrix}
^b\bm{R}_w & 0 \\
0 & ^b\bm{R}_w
\end{bmatrix} \cdot \bm{V}^{ref}_{ee}
\end{equation}
where $\bm{V}^{ref}_{ee}$ is the end-effector's velocity input w.r.t $T_w$.

Note that we choose Cartesian space velocity input $\bm{V}_a$ for the arm rather than $\dot{\bm{q}}_a$ in Joint space. It would be beneficial in handling the different dynamics of arm and base and the employment of CDMP, which will come to light in section \Rmnum{4}. The Cartesian velocity controller of the arm can be implemented either by analytical inverse kinematics ($\bm{V}_a \xrightarrow{\int} \bm{P}_a \xrightarrow{\text{IK}} \bm{q}_a$) with high accuracy \cite{craig2005introduction} or by Jacobian-based methods($\bm{V}_a \xrightarrow{\text{IK}} \dot{\bm{q}}_a \xrightarrow{\int} \bm{q}_a$) with singularity robustness 
\cite{nakamura_inverse_1986}.


\subsection{Weighted Whole-body Motion Distribution}
The kinematic redundancy of MCR, which brings motion flexibility, can generally be resolved by the following local optimization problem:
\begin{equation} 
\label{eqn:wln}
\underset{\bm{V}_{wb}}{\text{min}} \frac{1}{2}\|\bm{V}_{wb}\|_{\bm{W}} \quad ~ s.t.~ \bm{J}_{wb}\bm{V}_{wb} = \bm{V}_{ee}
\end{equation}
where $\bm{W} \in \mathbb{R}^{9\times9}$ is a symmetric and positive-definite weighting matrix. The general solution of $\bm{V}_{wb}$ to (\ref{eqn:wln}) can be obtained with:
\begin{equation} \label{eqn:wbsln}
\begin{aligned}
\bm{V}_{wb} &= \bm{J}^{+}_{wb}\bm{V}_{ee} + (\bm{I} - \bm{J}^{+}_{wb}\bm{J}_{wb})\bm{V}^0_n \\
\end{aligned}
\end{equation}
where $\bm{J}^{+}_{wb}=\bm{W}^{-1}\bm{J}^T_{wb}(\bm{J}_{wb}\bm{W}^{-1}\bm{J}^T_{wb})^{-1}$ is the weighted pseudo inverse of $\bm{J}_{wb}$, and $\bm{J}^{+}_{wb}\bm{V}_{ee}$ is the weighted least-norm(WLN) solution of (\ref{eqn:fk});\text{ }$(\bm{I} - \bm{J}^{+}_{wb}\bm{J}_{wb})\bm{V}^0_n \in \mathbb{N}(\bm{J}_{wb})$ is the homogeneous solution in the null space of $\bm{J}_{wb}$; $\bm{V}^0_n \in \mathbb{R}^9$ is an arbitrary vector that is usually used to improve some performance criterion $C(\bm{q}_{wb})$ without influencing the main task of end-effector's motion by replacing it with $k_n\nabla{C(\bm{q}_{wb})}$ in (\ref{eqn:wbsln}),
where $k_n$ is the control gain for the secondary task of optimizing $C(\bm{q}_{wb})$.

\subsection{Coupling Dynamic Movement Primitives}
DMP is a flexible method of encoding and generating trajectories, initially formulated by Ijspeert \etal\cite{ijspeert_movement_2002, ijspeert_dynamical_2013}.
The classical formulation of a discrete DMP for Cartesian position trajectories is given by the following dynamic system, which consists of a damped spring-like goal-attractor part that ensures convergence to the specified goal, and a nonlinear forcing term that allows the modeling of complex motion behavior:
\begin{equation}
\left\lbrace
	\begin{aligned}
		\tau\dot{\bm{z}} &= \alpha_{z}(\beta_{z}(\bm{g}-\bm{y})-\bm{z})+\bm{f}(s) \\
		\tau\dot{\bm{y}} &= \bm{z}
	\end{aligned}
\right.
\end{equation}
where $\tau, \alpha_z, \beta_z \in \mathbb{R}^+$, $\tau$ is a temporal scaling factor, $\alpha_z$ and $\beta_z$ define the convergence of the goal-attractor dynamics; $\bm{g}$, $\bm{z}$, $\bm{y} \in \mathbb{R}^3$ are respectively the goal, velocity and position of the system; $s$ is a phase variable monotonically evolving from 1 to 0 governed by the canonical system:
\begin{equation}
\tau\dot{s} = -\alpha_{s}s
\end{equation}
with $\alpha_s > 0$ and $s(0)=1$.
More details of DMP see \cite{ijspeert_dynamical_2013}.

As a reactive motion generator, DMP can be modulated during execution so as adapt to environmental uncertainties online by adding coupling terms \cite{saveriano_dynamic_2021}.
For instance, by incorporating an obstacle avoidance term $\bm{C}_o$  at \textit{acceleration} level \cite{hoffmann_biologically-inspired_2009}:
\begin{equation}
\tau\dot{\bm{z}} = \alpha_{z}(\beta_{z}(\bm{g}-\bm{y})-\bm{z})
+ \bm{C}_o
+\bm{f}(s)
\end{equation}
the encoded movement will be corrected to avoid obstacles.

To enable contact interaction with objects during reproduction, a DMP coupled with a force term $\bm{C}_f$ both at \textit{acceleration and velocity} level is  proposed in \cite{gams_coupling_2014}:
\begin{equation} \label{eqn:stiff-dmp}
\left\lbrace
	\begin{aligned}
		\tau\dot{\bm{z}} &= \alpha_{z}(\beta_{z}(\bm{g}-\bm{y})-\bm{z}) + k_d\dot{\bm{C}}_f +\bm{f}(s) \\
		\tau\dot{\bm{y}} &= \bm{z} + \tau\bm{C}_f \\
		\bm{C}_f &= k_p\bm{F}
	\end{aligned}
\right.
\end{equation}
where $k_p$ and $k_d$ are both scaling factors. $\bm{F}$ can be either a virtual defined or a real measured force. Applying the Laplace transform to the upper equation, we get transfer function from $\bm{F}$ to $\bm{y}$:
\begin{equation} \label{eqn:lp-admit}
\frac{\mathcal{L}(\bm{y})}{\mathcal{L}(\bm{F})}
= \frac{k_p(\tau^2+k_d)s+k_p\tau\alpha_z}{\tau^2s^2+\tau\alpha_z{s}+\alpha_z\beta_z}
\end{equation}
Compared with traditional admittance control scheme:
\begin{equation} \label{eqn:admit-ctrl}
\frac{\mathcal{L}(\bm{y})}{\mathcal{L}(\bm{F})}
= \frac{1}{{M}s^2 + {D}s + {K}}
\end{equation}
the dynamic behavior of (\ref{eqn:lp-admit}) is similar to  stiffness control of (\ref{eqn:admit-ctrl}) when $K \neq 0$ and has one more favorable derivative term of $\bm{F}$ in numerator. In this paper, we classify this type of force-coupled DMP as {\textit{Stiffness-coupled DMP}}.

In \cite{kramberger_passivity_2018}, the approach was further extended by adding the integral of the force coupling term to the goal $\bm{g}$, therefore exhibits similar behavior to the velocity resolved admittance control which defines $\dot{\bm{x}}_a = \bm{D}_a(\bm{F}_e - \bm{F}_d)$, where $\bm{x}_a$ is the position output, $\bm{D}_a$ is a symmetric positive-definite matrix, $\bm{F}_e$ and $\bm{F}_d$ are respectively external force and desired force. The formulation of this coupled DMP is:

\begin{equation} 
\label{eqn:admit-dmp}
\left\lbrace
	\begin{aligned}
		\tau\dot{\bm{z}} &= \alpha_{z}(\beta_{z}(\bm{g}+\bm{C}-\bm{y})-\bm{z}) +\bm{f}(s) \\
		\tau\dot{\bm{y}} &= \bm{z} + \tau\dot{\bm{C}} \\
		\dot{\bm{C}} &= \bm{D}_a(\bm{F}_e - \bm{F}_d)
	\end{aligned}
\right.
\end{equation}
Applying Laplace transform, we can obtain
\begin{equation} 
\label{eqn:tf-admit}
{\mathcal{L}(\bm{y})} = \frac{1}{s}\bm{D}_a{\mathcal{L}(\bm{F}_e)}
\end{equation}
which is equivalent to the velocity resolved admittance control, we classify it as {\textit{Admittance-coupled DMP}}.

\section{Methodology}
\subsection{Robust Whole-body Coordinated Control Scheme}
\begin{figure}[tb]
    \centering
    \includegraphics[width=1.0\linewidth]{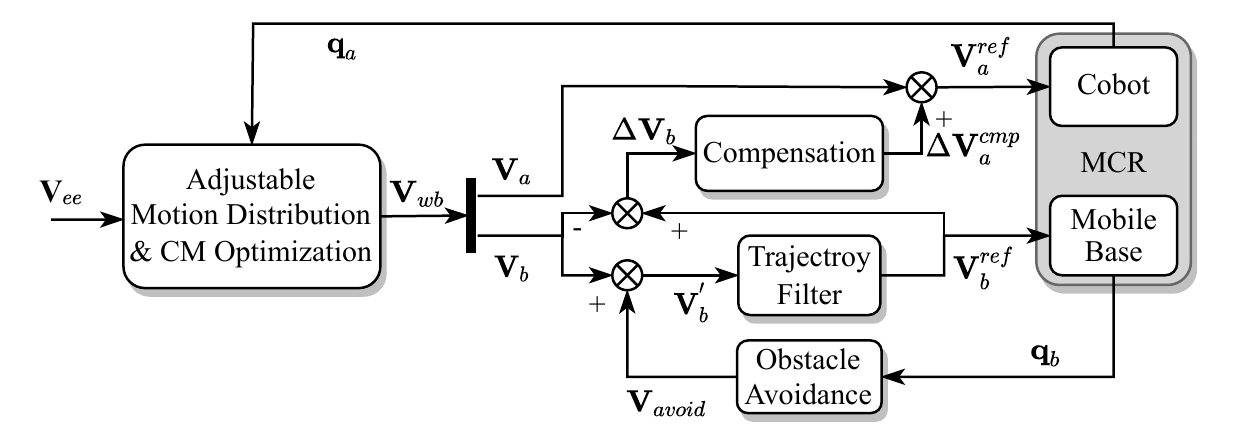}
    \vspace{-0.3cm}
    \caption{Block diagram of the proposed robust whole-body coordinated control}
    \vspace{-0.3cm}
    \label{fig:basic_wbc}
\end{figure}

As stated before, MCR bring better motion flexibility with wider working range and more degrees of freedom, but also lead to issues of coordination and dynamics coupling between the base and arm. So the design of whole-body control of MCR is threefold: first to optimize the capability of movement (CM) related to manipulability and joint limits through null-space movements, second to tackle the problem of different control bandwidth of base and arm to reduce structure vibration at the end-effector, and third to avoid constraints of both obstacles for safety and CM for robustness (i.e. to prevent the Cartesian pose controller failing when exceeding the constraint of CM). Fig.\ref{fig:basic_wbc} depicts the overall block diagram of the proposed robust whole-body coordination scheme.

\subsubsection{Adjustable Motion Distribution:} 
The weighting matrix $\bm{W}$ in (\ref{eqn:wln}) assigns the desired end-effector motion for the MCR. For better expression, we replace it by a new matrix 
\begin{equation}
\label{eqn:weight-factors}
\bm{Q}=\bm{W}^{-1}= 
\begin{bmatrix}
\bm{Q}_a & 0 \\
0 & \bm{Q}_b
\end{bmatrix}
\end{equation}
where 
$\bm{Q}_a = diag(\sigma_{xy}, \sigma_{xy}, 1, 1, 1, \sigma_{\phi})$, 
$\bm{Q}_b = diag(1-\sigma_{xy}, 1-\sigma_{xy}, 1-\sigma_{\phi})$.
$\sigma_{xy}, \sigma_{\phi} \in [0,1]$ are separately the weighting factors determining the motion distribution between base and arm in x-y plane and rotation around z-axis. $\sigma_{xy,\phi}=1$ and $\sigma_{xy,\phi}=0$ correspond to the sole motion of the arm and the base respectively, and $0<\sigma_{xy,\phi}<1$ means that the end-effector motion is achived by whole-body. Equation (\ref{eqn:weight-factors}) has a similar form proposed in \cite{jia_coordinated_2014} which uses a single factor for all Catesian motion. Here we present $\sigma_{\phi}$ specifically to distinguish rotation from translation which would be beneficial to pHRI applications.

\subsubsection{Movement Capability Optimization:} 
The manipulability index introduced by Yoshikawa \etal \cite{yoshikawa_manipulability_1985} is a quality measure of the distance to singularity for a given configuration of a manipulator. It can be computed as:
\begin{equation}
m = \sqrt{det(\bm{J}_a\bm{J}^T_a)} = \prod_{i=1}^{n}s_i
\end{equation}
where $s_i$ is the i-th singular value of $\bm{J}_a$. In \cite{Tsai_workspace_nodate}, the following penalization was introduced to consider joint limits:
\begin{equation}
P_{jnt} = 1-\exp(-k_{j}\prod_{i=1}^{n}\frac{(q_i-q_i^l)(q_i^u-q_i)}{(q_i^u-q_i^l)^2})
\end{equation}
where $k_j$ is a scaling factor adjusting the behavior near joint limits, $q_i^l$ and $q_i^u$ are the lower and upper limit of the i-th joint. We measure CM by the multiplication of $m$ with $P_{jnt}$:
\begin{equation}
C_m = m \cdot P_{jnt}
\end{equation} 

Substituting $C(\bm{q}_{wb})$ with $C_m$, Equation (\ref{eqn:wbsln}) is rewritten as:
\begin{equation} \label{eqn:wbccm}
\bm{V}_{wb} = \bm{J}^{+}_{wb}\bm{V}_{ee} + (\bm{I} - \bm{J}^{+}_{wb}\bm{J}_{wb})k_n
\begin{bmatrix}
	\nabla_x{C_m(\bm{q}_a)} \\
	\bm{0}_{3\times1}
\end{bmatrix}
\end{equation}
where $\nabla_x{C_m}$ is the gradient of CM w.r.t Cartesian pose coordinates of the arm. In practice, it is unnecessary to always optimize CM, $k_n$ can be set as zero when $C_m$ is greater than a certain threshold $C_m^u$.

\subsubsection{Trajectory Filtering:}
Generally, an MCR is composed of a heavy mobile base and a lightweight Cobot with considerable structure flexibility, therefore the base typically has slower dynamic response and lower bandwidth than the arm \cite{yamamoto_compensation_1994}. When an MCR is inputted velocity command with rapid change, the base can shake or even slip, which would cause strong vibration and tracking error at the end-effector due to the structure flexibility. Previous work try to solve the problem by distributing more motion to the arm adaptively \cite{jia_coordinated_2014,xing_enhancement_2020}, but the motion assigned to the base still contains high frequency motion. When both the base and arm use the Cartesian velocity control command as stated in Section \Rmnum{3}.A, it is a natural way to process motion commands of $\bm{V}_a$ and $\bm{V}_b$ separately and then compensate the error introduced by the process to ensure the end-effector's reference motion invariant.
We do not process $\bm{V}_a$ here as the arm can be viewed as an ideal actuator compared to the base due to its much higher bandwidth.
After obtained from Equation (\ref{eqn:wbccm}), $\bm{V}_b$ is corrected to $\bm{V}_{b}^{'}$ due to the module of obstacle avoidance and then is filtered out by a second-order low-pass filter:
\begin{equation}
\label{eqn:filtering}
\bm{V}_{b}^{flt}(s) = \frac{{\omega_f}^2}{s^2+2\zeta\omega_{f}s+{\omega_f}^2} \bm{V}_{b}^{'}(s)
\end{equation}
where $\omega_f$ is the cut-off frequency, $\zeta$ is the damping ratio, and $\bm{V}_{b}^{flt}$ is the filtered output which is sent to the base's velocity command interface as $\bm{V}_{b}^{ref}$.

\subsubsection{Robust Compensation and Constraint Avoidance:}
The deviation from $\bm{V}_{b}$ to $\bm{V}_{b}^{flt}$ should be compensated to keep the trajectory of the end-effector invariant, which is normally treated as top-priority task. However, there may exist conflict between the compensation and constraint-avoidance tasks. For example, collision avoidance may cause large motion deviation while the arm has reached its configuration constraint, thus it is required for the robot to deviate from the desired task motion. Hence it is necessary to cancel out the compensation motion in the direction of approaching the arm's configuration constraint.

The motion deviation of the base is:
\begin{equation}
\Delta \bm{V}_b = \bm{V}_{b}^{flt} - \bm{V}_{b}
\end{equation}
According to Equation (\ref{eqn:fk}), the required compensation motion of the arm can be computed by:
\begin{equation}
\Delta \bm{V}_a^{req} = -\bm{R}_{ba}^{T}\bm{J}_b \Delta\bm{V}_b
\end{equation}
To cancel out the motion worsenning the arm's CM, $\Delta \bm{V}_a^{req}$ is projected onto the orthogonal complement of $\nabla_x{C_m}$:
\begin{equation}
\label{eqn:wb_error}
\begin{aligned}
\Delta \bm{V}_a^{cmp} 
&=  [\bm{I} - \alpha \bm{P}_{C_m}] \Delta \bm{V}_a^{req} \\
&= [\bm{I} - \alpha \nabla_x^T{C_m}(\nabla_x{C_m} \cdot \nabla_x^T{C_m})^{-1}\nabla_x{C_m} ] \Delta \bm{V}_a^{req}
\end{aligned}
\end{equation}
Therefore, the reference velocity sent to the arm is:
\begin{equation}
\label{eqn:compensation}
    \bm{V}^{ref}_a = \bm{V}_a + \Delta\bm{V}_a^{cmp}
\end{equation}
Note that $\alpha$ is a transition function defined as:
\begin{equation}
\label{eqn:transition}
\alpha(C_m, C_m^{u}, C_m^{l})  = 
\begin{cases}
	0, & C_m \geq C_m^{u} \\
	0.5[1+cos(\displaystyle \frac{C_m-C_m^{l}}{C_m^{u}-C_m^{l}}\pi)], & C_m^{l} < C_m < C_m^{u} \\
	1,  & C_m \leq C_m^{l}
\end{cases}
\end{equation}
where $C_m^{u}$ and $C_m^{l}$ are respectively the upper and lower threshold of $C_m$. 

It is noteworthy that the ultimate motion tracking deviation of the end-effector is
\begin{equation}
\label{eqn:deviation}
\begin{aligned}
\Delta\bm{V}_{ee} 
&= \Delta \bm{V}_a^{cmp}-\Delta \bm{V}_a^{req}
&= - \alpha \bm{P}_{C_m} \Delta \bm{V}_a^{req}
\end{aligned}
\end{equation}

\subsection{Compliant Whole-body Control via CDMPs}
\begin{figure*}[tb]
    \centering
    \includegraphics[width=1.8\columnwidth]{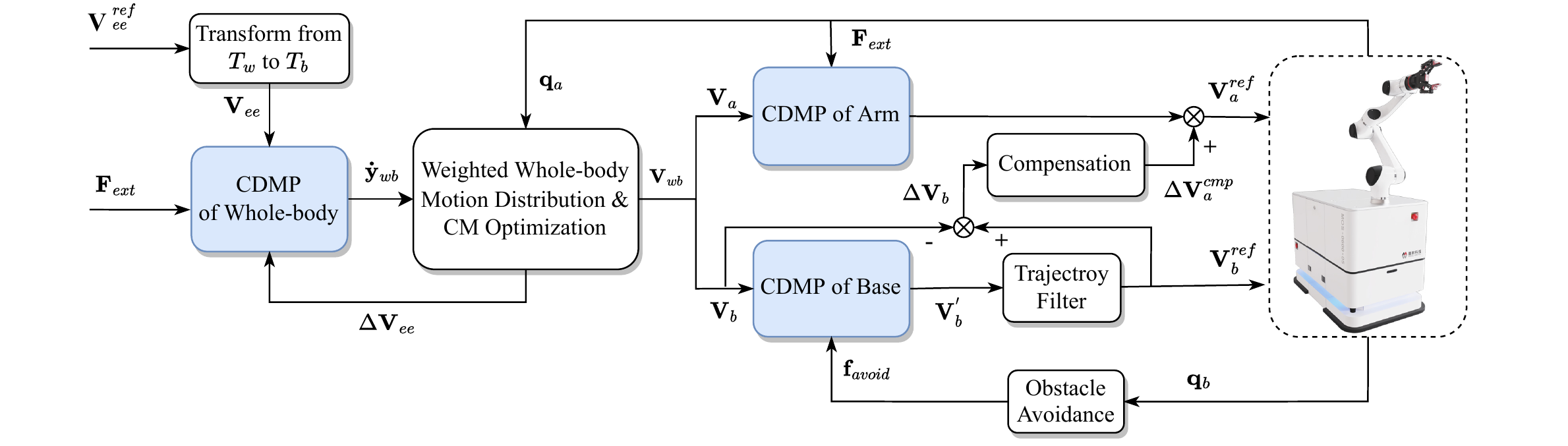}
    \caption{ The framework of CDMP-WBC. With interfaces of velocity control and force coupling, the designed CDMPs of whole-body, base, and arm are embedded into the whole-body coordinated control scheme introduced in Section \Rmnum{3}.A , and enable interactive and compliant functions.}
    \vspace{-0.8cm}
\label{fig:cdmp_wbc}
\end{figure*}
As shown in Equation (\ref{eqn:admit-dmp}), \textit{Admittance-coupled DMP} exibits equivalence as velocity control interface from $\dot{C}$ to $\dot{y}$. We can input velocity command into it by defining $\dot{\bm{C}}=\bm{V}+\bm{D}_a(\bm{F}_e - \bm{F}_d)$. Inspired by this, we design interfaces of both velocity control and force coupling for each DMP of the whole-body, base, and arm, respectively, and endow compliant abilities for the robot. Integrating with the designed DMPs, the unified whole-body control framework using CDMP is depicted as Fig. \ref{fig:cdmp_wbc}.

\subsubsection{Designed CDMPs for Whole-body, Base, and Arm:} 
To enable pHRI function for the MCR, an admittance interface term is added to the \textit{CDMP of Whole-body}:
\begin{equation} \label{eqn:cdmp-wb}
\left\lbrace
	\begin{aligned}
		\tau\dot{\bm{z}}_{wb} &= \alpha_{z}(\beta_{z}(\bm{g}_{wb}+\bm{C} - \bm{y}_{wb})-\bm{z}_{wb}) + \bm{f}(s) \\
		\tau\dot{\bm{y}}_{wb} &= \bm{z}_{wb} + \tau\dot{\bm{C}} \\
		\dot{\bm{C}} &= \bm{V}_{ee} + k_{hri} \bm{F}_{ext} \cdot S_{hri}
	\end{aligned}
\right.
\end{equation}
where $k_{hri}$ is the gain of admittance, $S_{hri}$ is a pHRI mode switching variable with value of 0 or 1, $\bm{F}_{ext}$ is the external force.

We fomulate the \textit{CDMP of Base} with a collision avoidance term as following:
\begin{equation} \label{eqn:cdmp-base}
\left\lbrace
	\begin{aligned}
		\tau\dot{\bm{z}}_b &= \alpha_{z}(\beta_{z}(\bm{g}_b+\bm{C} - \bm{y}_{b})-\bm{z}_b) + \bm{f}_{avoid}+\bm{f}(s) \\
		\tau\dot{\bm{y}}_b &= \bm{z}_b + \tau\dot{\bm{C}} \\
		\dot{\bm{C}} &= \bm{V}_b
	\end{aligned}
\right.
\end{equation}
where $\bm{f}_{avoid}$ is defined as $\nabla {U}_{rep}$, the gradient of a repulsive potential field function of
\begin{equation}
{U}_{rep}=
\begin{cases}
	0, & d_{obst} > d_{th} \\
	\frac{1}{2} k_{obs} (\frac{1}{d_{obs}}-\frac{1}{d_{th}})^{2}, & d_{obs} \leq d_{th}
\end{cases}
\end{equation}
where $d_{obs}$ is the distance to obstacles, $d_{th}$ is the distance threshold, and $k_{obs}$ is a scaling factor. 

As the arm has much higher tracking bandwidth and accuracy than the base, we add force control term to the \textit{CDMP of Arm} and propose a \textit{hybrid Stiffness/Force control coupled DMP} as follows:
\begin{equation}
\label{eqn:cdmp-arm}
\left\lbrace
	\begin{aligned}
		\tau\dot{\bm{z}}_a &= \alpha_{z}(\beta_{z}(\bm{g}_a+\bm{C}_{a} - \bm{y}_{a})-\bm{z}_a) + \bm{f}(s) + k_{d}\dot{\bm{C}}_{s} \\
		\tau\dot{\bm{y}}_a &= \bm{z}_a + \bm{C}_s + \tau\dot{\bm{C}}_{a}\\
		\bm{C}_s &= k_p \bm{S}_m \bm{F}_{ext} \\	
		\dot{\bm{C}_{a}} &= \bm{V}_{a} + k_f(\bm{I}-\bm{S}_m)(\bm{F}_d-\bm{F}_{ext})(1-S_{hri})
	\end{aligned}
\right.
\end{equation}
where $\bm{C}_s$ and $\bm{C}_a$ are respectively the stiffness-coupled term and admittance-coupled term, $k_p$ and $k_f$ are the respective gain variables. $\bm{F}_{d}$ is the desired force, $\bm{S}_m$ is a diagonal selected matrix with diagonal elements of 0 or 1. It will be force control when the element is set to be 0, stiffness control when 1. In addition, the force control term should be switched off when $S_{hri}$ is set to be 1 for scenario of pHRI.

\subsubsection{Eliminate Tracking Error by CDMP of Whole-body:} 
As mentioned above, the robust compensation strategy may cause tracking deviation at the end-effector. Utilizating the \textit{CDMP of Whole-body}, the deviation can be eliminated by adding the accumulated tracking error to $\bm{y}_{wb}$. The new formulation of \textit{CDMP of Whole-body} is as follows:
\begin{equation} \label{eqn:cdmp-wb-error}
\left\lbrace
	\begin{aligned}
		\tau\dot{\bm{z}}_{wb} &= \alpha_{z}(\beta_{z}(\bm{g}_{wb}+\bm{C} - (\bm{y}_{wb}+\int\Delta\bm{V}_{ee}dt))-\bm{z}_{wb}) + \bm{f}(s) \\
		\tau\dot{\bm{y}}_{wb} &= \bm{z}_{wb} + \tau\dot{\bm{C}} \\
		\dot{\bm{C}} &= \bm{V}_{ee} + k_{hri} \bm{F}_{ext} \cdot S_{hri}
	\end{aligned}
\right.
\end{equation}


\subsection{Intuitive Physical Human-Robot Interaction} 

\begin{figure}[tbp]
    \centering
    \includegraphics[width=0.47\linewidth]{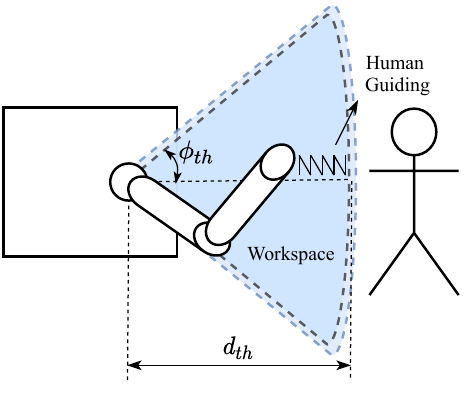}
    \vspace{-0.4cm}
    \caption{Local workspace of MCR with human workers.}
    \label{fig:local_ws}
\end{figure}

As stated before, the whole-body reference velocity from planning or human guidance can be distributed to the arm and base according to the corresponding weighting factors. Various motion modes can be achieved by adjusting the weighting factors in the way as shown in Table \ref{T:motion-mode}. 
\begin{table}[htbp]
 \vspace{-0.2cm}
\centering  
\caption{Various motion modes by the weighting factors}
\label{T:motion-mode}  
\begin{tabular}{p{2.5cm}<{\centering}m{3.2cm}<{\centering}m{0.7cm}<{\centering}m{1.8cm}<{\centering}m{0.7cm}<{\centering}}
\hline\noalign{\smallskip}
\textbf{Motion Mode} & \textbf{Weighting Factors}\\
\noalign{\smallskip}\hline\noalign{\smallskip}
      Locomotion     & $\sigma_{xy,\phi}=0$\\
      Manipulation   & $\sigma_{xy,\phi}=1$\\
      Loco-Manipulation  & $0<\sigma_{xy,\phi}<1$\\
      Intuitive pHRI & $0\leq \sigma_{xy,\phi}=f_{xy,\phi}(\bm{q}_a)\leq 1$\\
\noalign{\smallskip}\hline\noalign{\smallskip}
\end{tabular}
\end{table}
 \vspace{-0.4cm}

$f_{xy,\phi}(\bm{q}_a)$ are adaptive functions mapping the configuration of the arm to the weighting factors and defining the behavior of intuitive pHRI motion mode. In human-robot collaboration applications, this motion mode enables the base's fixedment and sole arm motion when working locally, and base's movement in case of distant working task or obstacle avoidance, which comply with the intuitiveness of operation.

\subsubsection{Local Workspace:} 
As shown as Fig. \ref{fig:local_ws}, the local workspace preferred by human workers is defined in a Cone-shape with two parameters $d_{th}$ and $\phi_{th}$, which are respectively the boundry of extend and deflection angle of the end-effector w.r.t to the arm's base. The CM of the arm is penalized when entering the boundries $[d^l, d^u]$ and $[\phi^l, \phi^u]$ by penalization factors $P_{d} = \alpha(d, d^u, d^l)$ and $P_{\phi} = \alpha(\phi, \phi^u, \phi^l)$, where $\alpha$ is the transition function defined by Equation (\ref{eqn:transition}) to ensure smooth transition. Defining $C_m^d = C_m \cdot P_{d} \cdot P_{\phi}$, $C_m^{\phi} = C_m \cdot P_{\phi}$, $f_{xy,\phi}$ are then calculated as
\begin{equation}
\left\lbrace
	\begin{aligned}
		f_{xy} &=  1 - \alpha(C_m^d, C_m^{u}, C_m^{l})\\
		f_{\phi}&= 1 - \alpha(C_m^{\phi}, C_m^{u}, C_m^{l})
	\end{aligned}
\right.
\end{equation}
which will be 0 when exceeding the workspace, 1 when staying in the workspace, and between 0 and 1 when crossing the boundaries.

\subsubsection{Deactivation Strategy:} 
When the human worker stretches the arm to a configuration in singularity or out of the local workspace, the base moves solely. Deactivation of this state is needed when the motion approaches away from singularity or towards the local workspace, otherwise the robot would be locked at such configuration. To this end, the following deactivation strategy is designed:
\begin{equation}
\sigma_{xy, \phi} = 
\begin{cases}
	1,  &  \text{if}~(\nabla_xC_m^{d,\phi} \cdot \bm{V}_{ee} > 0) \vee ({\dot{d},\dot{\phi}}<0) \\
	f_{xy,\phi}, & \text{otherwise}
\end{cases}
\end{equation}

\section{Experiments}
Several real-world experiments were carried out with a velocity-controlled MCR to demonstrate the effectiveness of our proposed controller. The documented video of these experiments can be found at \href{https://youtu.be/VVWfm8\_f8QA}{https://youtu.be/VVWfm8\_f8QA}.

\subsection{Experimental Setup}
The MCR used for experiments consists of a Moying's 3-DoF omnidirectional mobile base and a 6-DoF Elfin5 industrial collaborative manipulator equipped with a Robotiq FT-300 F/T sensor on the end-effector. Two 2-D LIDAR sensors (TiM571-9950101S01) are mounted on the diagonally opposite of the base. The pose of the base can be estimated by the internal odometry, and the pose of the end-effector is measured by joint encoders. An OptiTrack indoor motion capture system is also used in the experiment of trajectory tracking to measure the real end-effector pose with structure vibration. 
The proposed controller is implemented based on ROS. It is run on an industrial computer with four Intel Core i5-8500 CPUs and 16 GB of RAM. The sampling rate of the F/T sensor and the laser radars is 100 Hz and 15 Hz, respectively. The control rate is set to be 100 Hz which is the maximum sampling frequency of the used Robotiq FT-300 F/T sensor.
\subsection{Trajectory Tracking}
\begin{figure}[htb]
    \centering
    \includegraphics[width=1.03\linewidth]{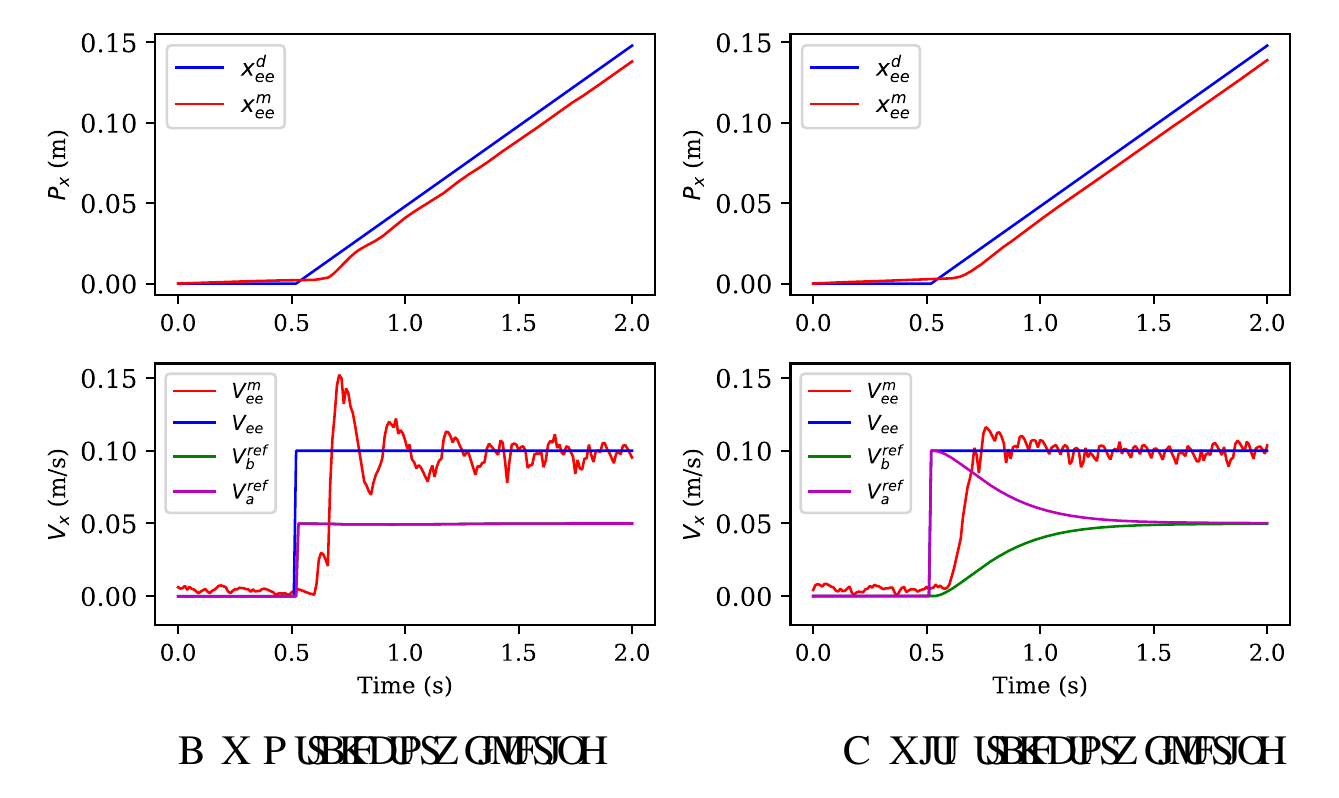}
    \vspace{-0.7cm}
    \caption{Experimental results of trajectory tracking.}
    \vspace{-0.4cm}
    \label{fig:traj_opt}
\end{figure}

To test the effectiveness of the proposed whole-body coordinated controller in addressing the issue of different dynamics of base and arm, a trajectory tracking experiment was performed. A step velocity command of 0.1 m/s along the x-axis of $T_b$ was sent to the end-effector to excite vibration. Two sets of experiments were executed: one with trajectory filtering($\omega_f=2\pi$ rad/s, $\zeta=1.0$) as described in the method; one without trajectory filtering for comparison. $\sigma_{xy}$ is set to be 0.5, so $\bm{V}_{ee}$ is equally distributed into $\bm{V}_b$ and $\bm{V}_a$ by Equation (\ref{eqn:wbccm}), then $\bm{V}_b$ is filtered into $\bm{V}^{ref}_b$ by (\ref{eqn:filtering}), and $\bm{V}_a$ is compensated into $\bm{V}^{ref}_a$ by Equation (\ref{eqn:compensation}). The experimental results are shown by Fig. \ref{fig:traj_opt}.

The end-effector's position was measured by OptiTrack at 100 Hz, and is filtered through a low-pass filter with cut-off frequency of 40 rad/s to compute the velocity through difference calculation. Fig. \ref{fig:traj_opt} shows that $\bm{V}^{ref}_b$ transits smoothly with trajectory filtering and the oscillation of the end-effector reduced significantly than without trajectory filtering.

\subsection{Robust Whole-body Control}

\begin{figure}[htp]
    \centering
    \includegraphics[width=\linewidth]{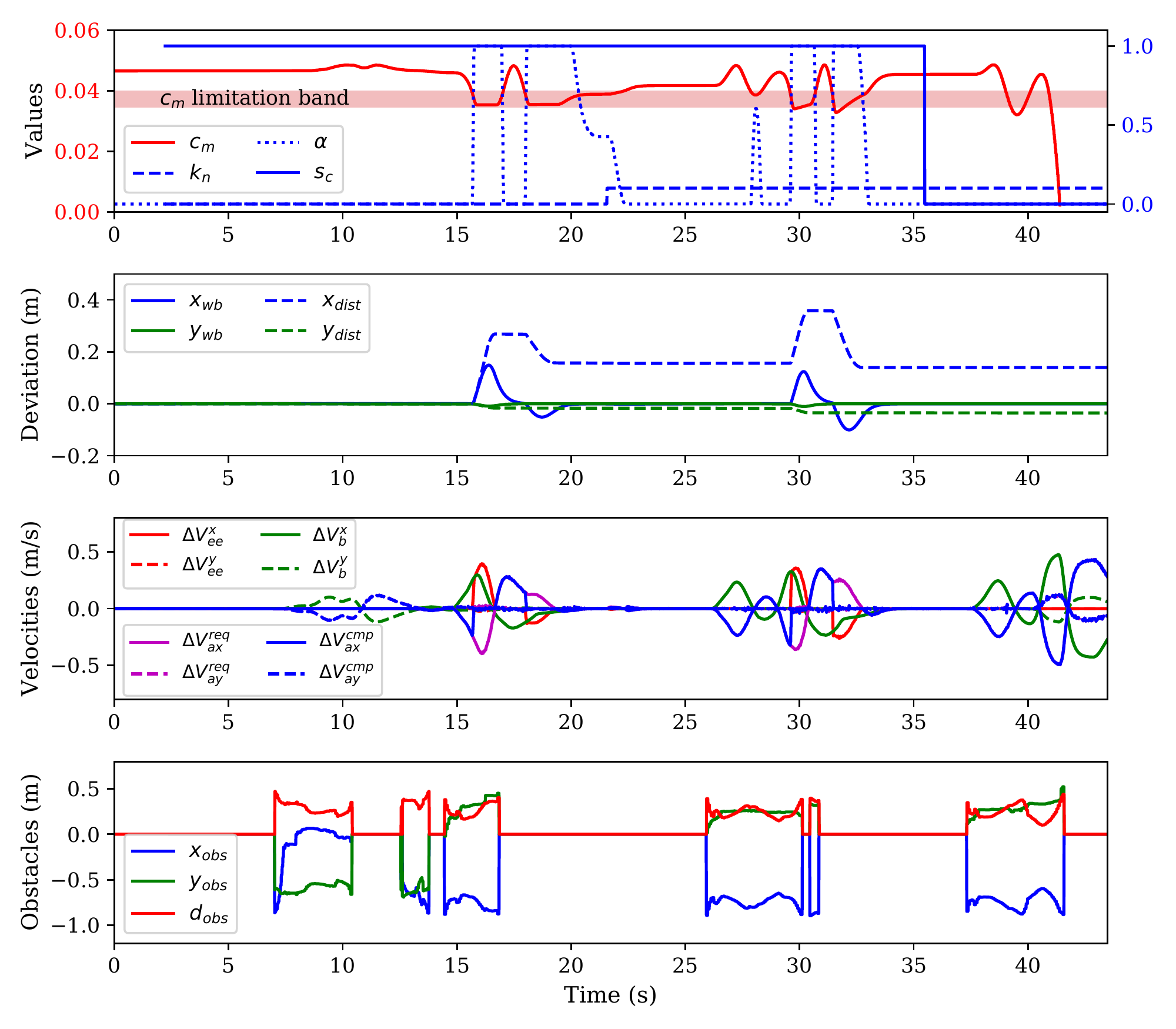}
\vspace{-0.5cm}
    \caption{Fixed positioning task while avoiding obstacles. Note that $\Delta V^{cmp}_a$ cannot compensate $\Delta V_{b}$ well anymore when CM decreases into the limitation band, which results in the motion traction deviation $\Delta V_{ee}$ as described by Equation (\ref{eqn:deviation}).}
    \label{fig:wbc_avoid}
\end{figure}

Fig. \ref{fig:wbc_avoid} shows the result of the experiment of obstacle avoidance, which was performed to test the proposed robust whole-body coordination and compensation strategy. $x_{dist}$ and $y_{dist}$ indicate the perturbed deviation $\int\Delta\bm{V}_{ee}dt$ in the x and y directions as described by Equation (\ref{eqn:deviation}). $d_{th}=0.5$ m and $k_{obs}=0.01$.
A person approached four times to the base with different settings while the end-effector was commanded to stay still. At the first time, the optimization of CM was switch off ($k_n=0$), and the compensation strategy was switched on ($s_c=1$) as the arm was away from configuration constraints. From the velocity splines we can see that the base avoided the person, and the arm compensated the deviation well, assuring the tracking of $x_{wb}$. At the second time, the person approached the base closer, and CM decreased into the limitation band ($0.035<C_m<0.04$). Thanks to the robust compensation strategy of Equation (\ref{eqn:wb_error}), CM did not decrease anymore at the lower limit because of the cancellation of the compensation motion in the direction of deteriorating CM($0<\alpha\leq1$). However, due to the cancellation, the end-effector deviated from the commanded fixed point, but then returned after the person moved away thanks to the tracking error elimination strategy of Equation (\ref{eqn:cdmp-wb-error}). Then at the time of about 22 s, $k_n$ is activated from 0 to 0.1, obtaining a consequent increase of CM through nullspace movements until the limitation band is exceeded, while still successfully tracking $x_{wb}$.

\vspace{-0.2cm}

At the third time, when the person approached the base even closer than the second time, CM did not decrease anymore when reaching the lower boundary of the limitation band and even had a tendency to increase as the optimization of CM was working ($k_n=0.1$).
At the last time, at the time of 35.5 s, the robust compensation strategy is manully switched off ($s_c=0$ and $\alpha=0$ afterwards), when the person approached the base again, the arm compensated base's deviation exactly ($\Delta V_{ee}=0$) yet CM decreased rapidly because of the deteriorating compensation motion until the Cartesian velocity controller failed.

\begin{figure}[htbp]
    \centering
    \vspace{-0.3cm}
    \includegraphics[width=0.85\linewidth]{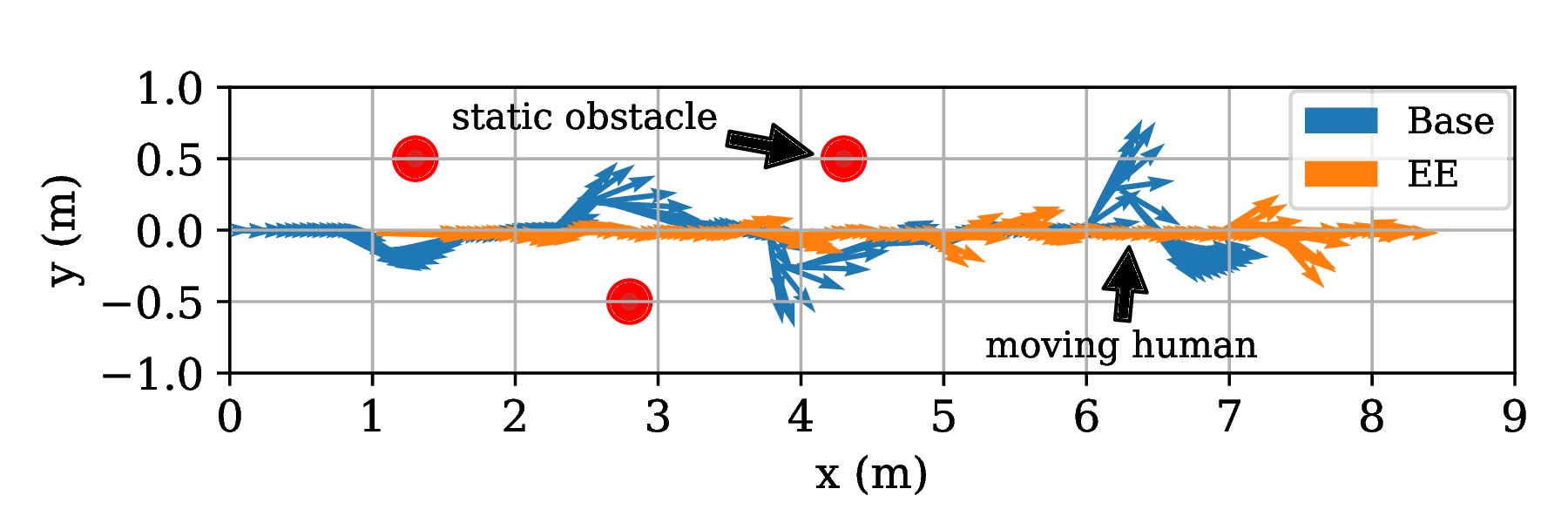}
    \vspace{-0.3cm}
\caption{Vector histories of base and end-effector while tracking a line through static and moving obstacles}
    \label{fig:obst_avoid}
\end{figure}
 \vspace{-0.3cm}

Fig. \ref{fig:obst_avoid} shows the history of position-velocity vectors of the base and arm's end-effector during tracking a line at the speed of 0.3 m/s through static obstacles and moving humans. It is noticeably shown that the motion deviation of the end-effector is much lower than the base's, and fairly maintain a straight line, which validates the effectiveness of the compensation strategy.

\subsection{Force Control for Wiping Task}
\begin{figure}[htbp]
    \centering
    \includegraphics[width=1.0\linewidth]{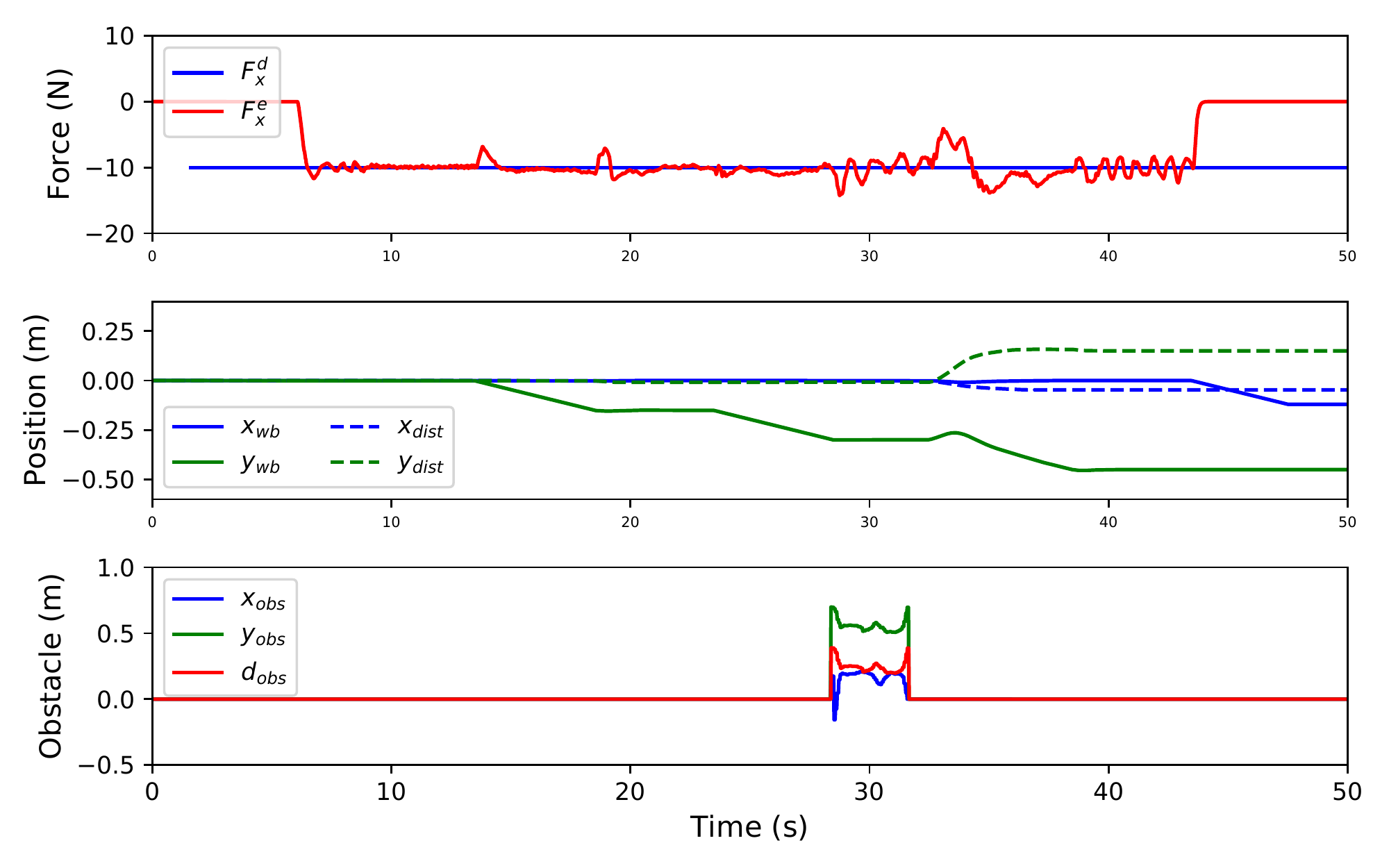}
    \vspace{-0.5cm}
    \caption{Wiping task while avoiding obstacles. $F^d_x$ and $F^e_x$ are the desired and the measured external force in x direction respectively. $x_{dist}$ and $y_{dist}$ represent the perturbed motion deviation.}
    \label{fig:wipe}
\end{figure}

Fig. \ref{fig:wipe} shows the result of wiping task while avoiding obstacles. The desired force was set to be -10 N, the gain of force control $k_{f}=0.01$, and stiffness control $k_p=0.0005$. The selected matrix of hybrid Stiffness/Force control was set to be $\bm{S}_m=diag(0,1,1,1,1,1)$, which means that it was force control in x direction and stiffness control in other directions to execute wiping motion. It can be seen that the force tracking worked well while wiping even when a person approached the base at the time of about 28 s.

    \vspace{-0.5cm}

\subsection{Intuitive Human-Robot Interaction}
\begin{figure*}[tp]
    \centering
    \includegraphics[width=1.42\columnwidth]{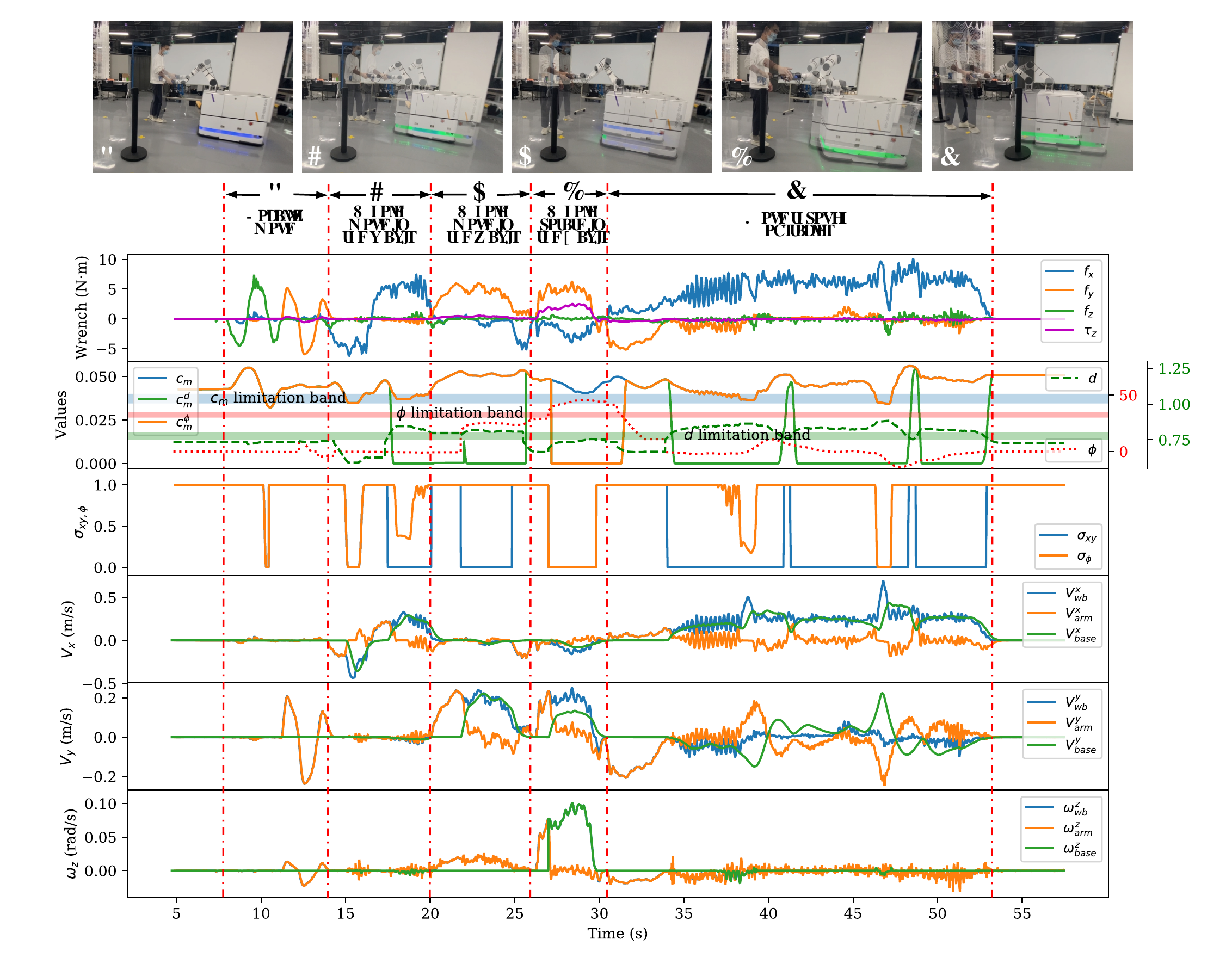}
    \vspace{-0.3cm}
    \caption{Intuitive physical human-robot interaction}
\label{fig:hri_pic}
\vspace{-0.8cm}
\end{figure*}

Fig. \ref{fig:hri_pic} shows the result of pHRI experiment. The $S_{hri}$ was activated as 1, and $k_{hri}=0.05$. During A stage, the arm moved solely as the end-effector was staying in the local workspace; during B and C stages, the base began to move in the direction of x and y respectively as the $C_m$ decreased to the boundary ($d^l=0.75$ m, $d^u=0.8$ m). At the stage of D, the deflection angle exceeded the boundary ($\phi^l=30^{\circ},\phi^u=35^{\circ}$), the base began to rotate as $\sigma_{phi}$ was lower than 1. At the stage of E, the whole robot was dragged through an obstacles zone. It can be seen that the base avoided the obstacles successfully and the pHRI process complied with human's habits.

\section{Conclusion}
\label{sec:conclusion}


This paper presents a whole-body control framework using CDMP-WBC for velocity-controlled MCR. The framework consists of a robust whole-body coordinated scheme, which tackles the different dynamics of base and arm and manage the avoidance of various constraints in a multi-hierarchy prioritized way. Compliant whole-body control is achieved through a proposed novel CDMP-based scheme. Also, intuitive pHRI is realized in this unified framework.
Hardware experiments validated the effectiveness of the proposed method. The method can be universally applied to velocity-controlled MCR which are more common and low-cost in industrial applications compared with torque-controlled ones. Note that the nonlinear forcing term $\bm{f}(s)$ is set to be zero in all the experiments. Future work will focus on further leveraging the learning and planning abilities of DMP by taking advantage of $\bm{f}(s)$.
\bibliographystyle{ieeetr}
\bibliography{ref.bib}

\end{document}